\documentclass{Interspeech}

\interspeechcameraready 

\newcommand{\methodsname}{MAS-LoRA}
\newcommand{\fullmethodsname}{Mixture of Accent-Specific LoRAs}
\usepackage[inkscapeformat=png]{svg}
\usepackage{multirow}
\usepackage{caption}
\usepackage{subcaption}
\usepackage{array,etoolbox}
\preto\tabular{\setcounter{magicrownumbers}{0}}
\newcounter{magicrownumbers}
\def\rownumber{}

\title{Mixture of LoRA Experts for Low-Resourced Multi-Accent\\ Automatic Speech Recognition}

\author[affiliation={1}]{Raphaël}{Bagat}
\author[affiliation={1}]{Irina}{Illina}
\author[affiliation={1}]{Emmanuel}{Vincent}

\affiliation{Université de Lorraine, CNRS, Inria, LORIA}{F-54000 Nancy}{France}
\email{raphael.bagat@loria.fr, irina.illina@loria.fr, emmanuel.vincent@inria.fr}

\usepackage{comment}

\usepackage{pgfplots}
\pgfplotsset{compat=1.9}
\usepgfplotslibrary{external}
\begin{document}

\maketitle

\begin{abstract}

We aim to improve the robustness of Automatic Speech Recognition (ASR) systems against non-native speech, particularly in low-resourced  multi-accent settings. 
We introduce \fullmethodsname{} (\methodsname{}), a fine-tuning method that leverages a mixture of Low-Rank Adaptation (LoRA) experts, each specialized in a specific accent.
This method can be used when the accent is known or unknown at inference time, without the need to fine-tune the model again.
Our experiments, conducted using Whisper on the L2-ARCTIC corpus, demonstrate significant improvements in Word Error Rate compared to regular LoRA and full fine-tuning when the accent is unknown.
When the accent is known, the results further improve.
Furthermore, \methodsname{} shows less catastrophic forgetting than the other fine-tuning methods.
To the best of our knowledge, this is the first use of a mixture of LoRA experts for non-native multi-accent ASR.

\end{abstract}

\section{Introduction}

Automatic Speech Recognition (ASR) systems have reached human-like performance in many domains \cite{xiong2017towards}.
End-to-end systems such as Whisper \cite{radford2022robust}, a multilingual ASR model, work very well when the speakers talk in their native language.
However,
their performance drops on non-native, accented speech. 
Indeed, non-native speech often involves specific pronunciations of certain phonemes borrowed from the speaker's mother tongue (L1) \cite{zampini2008l2}, which induce ASR errors.
Non-native accent can also affect the prosody of the utterance to resemble the speaker's L1, leading to an even greater mismatch with native speech \cite{busa2010effects}.
In the context of \textit{multi-accent} ASR, when the systems are used to transcribe utterances from different accents, these phenomena are exacerbated by the larger number of accents.
When facing accented speech, ASR systems can either be \textit{accent-agnostic}, i.e. have no information about the speaker's accent, or on the contrary, be \textit{accent-aware}.
For a system to be used in an accent-agnostic setting, transcribed training data that cover a wide variety of accents are needed.
Such data are rare, thus ASR systems must be trained on low-resourced data which makes the problem even more challenging.
Improvements in non-native multi-accent ASR would make these systems usable in contexts where people have to speak a different language than their mother tongue, e.g., in Air Traffic Communications where pilots from all over the world have to speak English, or in international commerce.

Initial approaches explored the adaptation of Gaussian mixture model - hidden Markov model (GMM-HMM) based acoustic models for accented ASR in both accent-aware and accent-agnostic settings \cite{vergyri2010automatic,kamper2011multi}.
More recently, deep learning based models have been studied to improve accented ASR.
Especially in the case of multi-accent ASR, prior works proposed to improve ASR by using accent recognition in a multi-task setting to learn accent specific features along the ASR training \cite{jain2018improved,viglino2019end}.
Methods based on adding one-hot representations of dialects to the model's input also showed promising improvements \cite{li2018multi}. However, these methods considered native accents only. 

To bridge the gap with non-native accents, \cite{matassoni2018non} used various transfer learning methods to improve non-native multi-accent ASR, exhibiting the importance of a multilingual model to handle pronunciation differences across accents.
This method is based on full fine-tuning, which is computationally expensive. Parameter-efficient fine-tuning methods have emerged,  starting with Adapters \cite{pmlr-v97-houlsby19a} which consist of training small neural modules inserted in between a model's pre-existing layers while keeping these layers frozen.
\cite{huangadapter} used Adapters to fine-tune Whisper with different native English accents, leading to similar results to full fine-tuning and even improvements for the African-American accent.
\cite{gong2022layer} used multiple (non-linear) Adapters to improve non-native multi-accent ASR, but this method relies on an external accent identification model.
Following Adapters, \textit{Low-Rank Adaptation} (LoRA) \cite{hu2022lora} and its many variants \cite{zhang2023adalora,hayou2024lora+,liu2024dora} have been proposed to further improve parameter-efficient fine-tuning.
LoRA has been used to improve ASR systems on specific languages \cite{li2025review}.
In order to use LoRA on data coming from different domains, many methods proposed to jointly use multiple LoRAs as a mixture of experts (MoE) \cite{luo2024moelora,gao2024higher,li2024mixlora} and \cite{song2024lora} use them to improve Whisper's multilingual ASR.
To the best of our knowledge, mixture of LoRA expert methods have not yet been used for non-native multi-accent ASR, which remains an understudied problem due in particular to its low-resourced nature.

\begin{figure*}[th]
  \centering
  \begin{subfigure}{0.335\textwidth}
      \includegraphics[width=\linewidth]{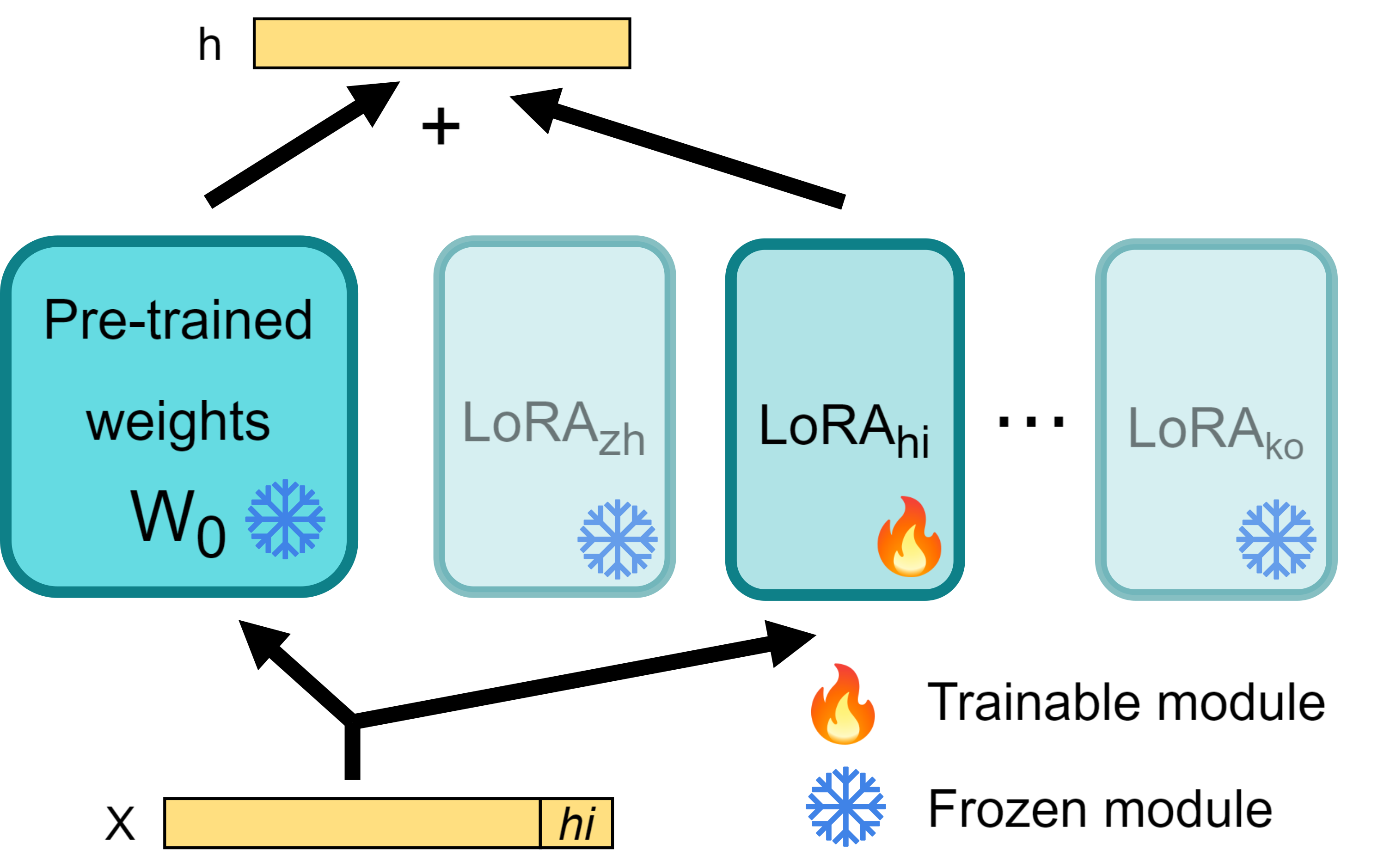}
      \caption{Architecture of \methodsname{} during training, here, a sample of accent "hindi" is being processed.}
      \label{fig:aslora_training}
  \end{subfigure}
  \begin{subfigure}{0.335\textwidth}
      \includegraphics[width=\linewidth]{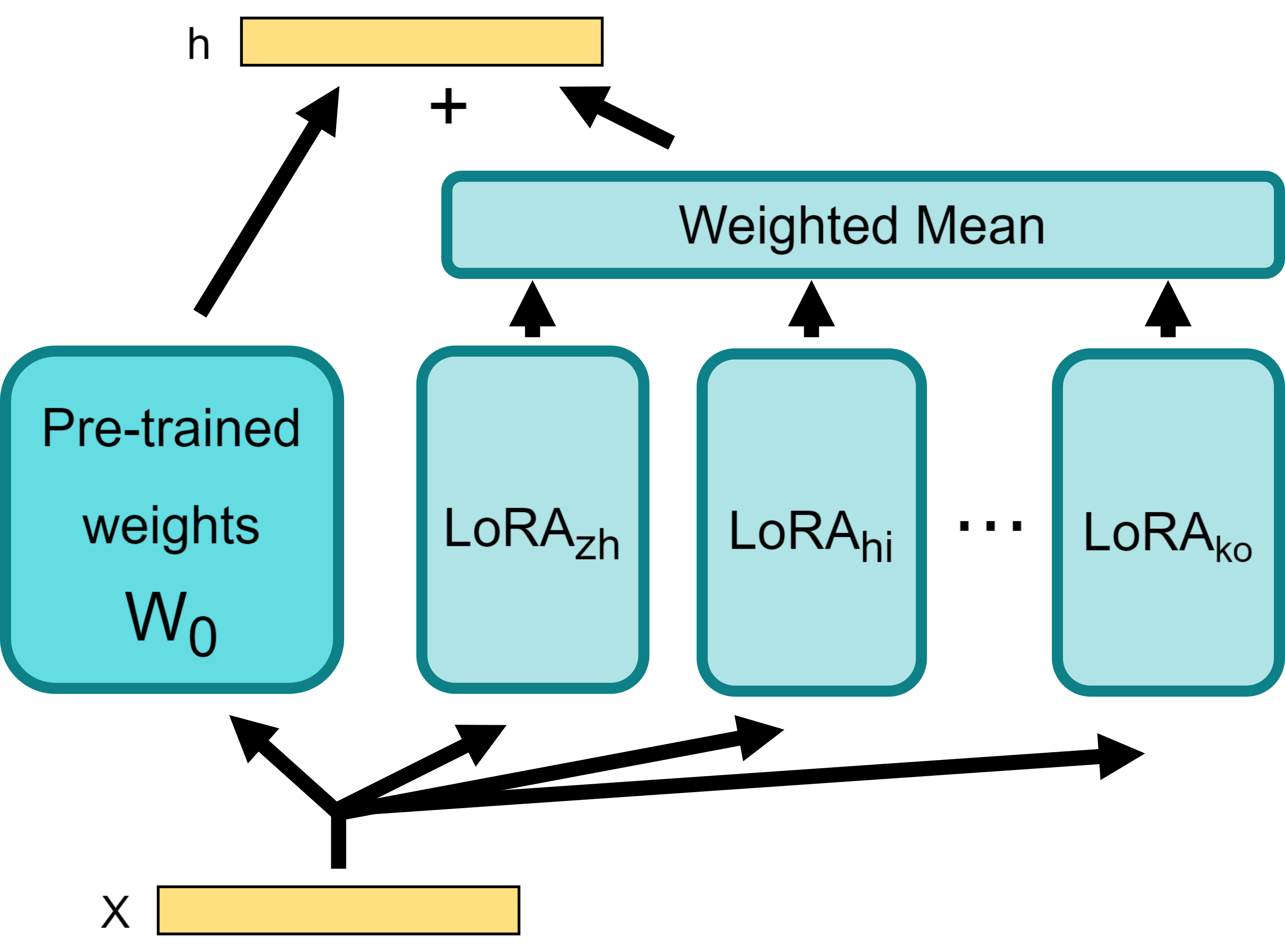}
      \caption{Architecture of \methodsname{} during inference.}
      \label{fig:aslora_inference}
  \end{subfigure}
  \caption{Architecture of \methodsname{}.}
  \label{fig:aslora_fig}
\end{figure*}

In this paper, we leverage Whisper's multilingual knowledge via a mixture of LoRA experts to improve non-native multi-accent ASR.
Each expert specializes in a single accent and their combined knowledge is used at inference time, either with equal weights for all accents in an accent-agnostic setting or with a fixed, higher weight for the target accent and a lower equal weight for the remaining accents in an accent-aware setting. We show that both approaches decrease the WER on the L2-ARCTIC corpus compared to using a single LoRA corresponding to the ground truth accent, which validates the MoE approach. Thanks to the linear nature of LoRA, the weights of the LoRA experts can be merged with those of the original model, leading to non-native multi-accent ASR at no extra computational cost.

This paper is organized as follows. 
Section \ref{sec:proposed_method} introduces the proposed method. 
Section \ref{sec:data_methodo} presents our experiments.
Section~\ref{sec:results_discussions} describes our results. We conclude in Section \ref{sec:conclusion}.

\section{Proposed methodology}\label{sec:proposed_method}
\subsection{Classical LoRA}
LoRA aims to approximate weight updates $\Delta W \in \mathbb{R}^{d \times k}$ of the frozen pre-trained weights $W_0 \in \mathbb{R}^{d \times k}$ during fine-tuning by the product of two low-rank matrices $A \in \mathbb{R}^{r \times k}$ and $B \in \mathbb{R}^{d \times r}$ with rank $r \ll \min(d,k)$ scaled by a factor $\alpha$:
\begin{align}
    W &= W_0 + \alpha\, \Delta W \nonumber \\
      &= W_0 + \alpha\, BA.
\label{eq:lora}
\end{align}
While LoRA can be applied to any types of pre-trained weights, it is extensively used in attention modules by applying it to some or all of the attention matrices.
By construction, its weights can be merged with the pre-trained weights and do not increase the computational cost at inference time.

\subsection{\fullmethodsname{} (\methodsname{})}
We propose \methodsname, a MoE method using LoRA experts trained on single-accent data and combined at inference time to process multi-accent data.
In detail, if the training data contains \textit{n} accents, we instantiate $n$ LoRA experts, one per accent, which were used at the same time as $W_0$.
This way, each expert is adapted to a specific accent, making it easier to learn each accent's unique characteristics.

\subsubsection{Accent-specific fine-tuning}
During fine-tuning, samples with a given accent will only pass through the pre-trained weights and the expert specialized in this accent (see Fig.~\ref{fig:aslora_training}).
As with LoRA, the experts use low-rank-parametrized update matrices. For a sample of hidden representation $x$ and accent $j$, the output $h$ of one \methodsname{} layer during training is
\begin{align}
    h &= \text{\methodsname{}}_{j}(x) \nonumber \\
      &= (W_0 + \alpha\, \Delta W_{j})\,x \nonumber \\
      &= (W_0 + \alpha\, B_{j}A_{j})\,x.
\label{eq:aslora_training}
\end{align}
This allows each expert to be trained separately.
Similarly to other parameter-efficient fine-tuning methods, the pre-trained weights $W_0$ remain frozen throughout the entire fine-tuning.
As shown in \cite{gong2022layer}, in an encoder-decoder architecture, accent-related adaptation for the encoder leads to a systematic improvement.
Thus, we chose to always use \methodsname{} to fine-tune the encoder.
Its use in the decoder is discussed in Section \ref{sec:where_to_apply}.


As opposed to regular MoEs, we do not learn routers, as we shall see in Section~\ref{sec:results_discussions} that assigning all weight to the expert corresponding to the ground truth accent, what we could make routers learn, is suboptimal w.r.t. sharing weight with other experts.
We have tried learning routers, but it did not show any improvement.

\begin{table*}[th]
  \caption{WER (\%) obtained with different fine-tuning methods on L2-ARCTIC and LibriSpeech test-clean. The Encoder and Decoder columns indicate the fine-tuning method used in the encoder and the decoder. The percentage of trained parameters is with respect to the total model size. Bold numbers indicate the best result for each corpus and those results which are statistically equivalent to it.}
  \label{tab:results_big}
  \centering
  \begin{tabular}{@{\makebox[3em][r]{\rownumber\space}} |c c|c|c c }
    \toprule
    {\textbf{Encoder}} & {\textbf{Decoder}}  & {\textbf{Trained params. (\%)}} & {\textbf{WER L2-ARCTIC (\%)}} & {\textbf{WER LibriSpeech (\%)}}  \gdef\rownumber{\stepcounter{magicrownumbers}\arabic{magicrownumbers}} \\
    \midrule
    \midrule
    No FT & No FT &  0 & 13.77 & \textbf{5.78} \\
    Full FT & Full FT & 100 & 12.21 & 7.90 \\
    \midrule
    \midrule
    LoRA-qv & LoRA-qv & 0.73 & 12.32 & 6.32 \\
    \midrule
    \multirow{3}{*}{\methodsname{}-qv} & No FT & 1.44 & 14.08 & \textbf{5.94} \\
      & LoRA-qv & 1.91 & \textbf{11.77} & \textbf{5.81} \\
      & \methodsname{}-qv & 4.21 & \textbf{11.78} & \textbf{5.91} \\
    \midrule
    \midrule
    LoRA-qkvo & LoRA-qkvo & 1.44 & 13.48 & 7.16 \\
    \midrule
    \multirow{3}{*}{\methodsname{}-qkvo} & No FT & 2.84 & 12.14 & \textbf{5.95} \\
      & LoRA-qkvo & 3.76 & \textbf{11.77} & \textbf{5.95} \\
      & \methodsname{}-qkvo & 8.07 & \textbf{11.90} & 6.27 \\
    \bottomrule
  \end{tabular}
\end{table*}

\subsubsection{Accent-agnostic inference}
At inference (see Fig.~\ref{fig:aslora_inference}), when the accent of the sample is unknown, we average the outputs of all experts $W_i$ with equal $\frac{1}{n}$ weights before adding them to the output of the pre-trained weights $W_0$:
\begin{align}
    W &= W_0 + \frac{1}{n}\sum^n_{i=1} \alpha\, W_i \nonumber \\
      &= W_0 + \frac{1}{n}\sum^n_{i=1} \alpha\, B_iA_i.
\label{eq:aslora_inference}
\end{align}
This mixture method allows us to merge experts with pre-trained weights, preserving the original inference cost.

\subsubsection{Accent-aware inference}
When the accent label is available at inference time, instead of using an equal weight of $\frac{1}{n}$ for every accent, it is possible to give a higher weight to the expert corresponding to that accent.
We parameterize that weight as $\frac{1}{\beta}$, with $\beta \in [1,n]$. The residual $1-\frac{1}{\beta}$ weight is shared equally among all other accents.
Denoting as $j$ the accent label of the current sample, the experts are used as follows:
\begin{align}
    W &= W_0 + \frac{1}{\beta}\,\alpha\,B_{j}A_{j} + \frac{1-\frac{1}{\beta}}{n-1} \sum^n_{\substack{i=1\\i\neq j}} \alpha\,B_iA_i.
\label{eq:aslora_beta}
\end{align}
We do not merge experts, causing only a slight increase in inference cost.

\section{Experimental settings} \label{sec:data_methodo}
\subsection{Datasets}
Our experiments are conducted on the L2-ARCTIC dataset \cite{zhao2018l2}. 
This dataset contains speech utterances in English spoken by non-native speakers with different accents.
The accents (L1) are the following: \textit{Arabic}, \textit{Chinese}, \textit{Hindi}, \textit{Korean}, \textit{Spanish}, and \textit{Vietnamese}.
Each accent class has 4 different speakers, thus totaling 24 speakers, with 1~h of data per speaker.
Every speaker reads the same phonetically-balanced sentences originating from Project Gutenberg \cite{kominek2004cmu}.

To avoid evaluation biases, it is important that the sentences and speakers in the test set are disjoint from those in the training and validation sets. Ideally, the sentences in the training and validation sets should also be disjoint.
Due to the small amount of data, we run 8-fold cross-validation. For a given fold and accent, the training set contains 80\% of the unique sentences spoken by 3 speakers, the validation set contains 10\% other sentences spoken by the same 3 speakers, and the test set contains the 10\% remaining sentences spoken by the remaining speaker. Thus, each speaker is part of two test folds. Each method is fine-tuned and tested using the same 8 folds.
Table \ref{tab:data} shows the split for a single accent across all folds.

\begin{table}[th]
  \caption{Quantity of audio and words per accent across all folds. Each accent follows the same split.}
  \label{tab:data}
  \centering
  \begin{tabular}{ c | c c c }
    \toprule
    & {\textbf{Training}} & {\textbf{Valid.}} & {\textbf{Test}}   \\
    \midrule
    \textbf{Audio duration} & {8x 2~h 48~min} & {8x 18~min} & {8x 6~min} \\
    \midrule
    \textbf{\# of words} & {8x 144,154} & {8x 17,028} & {8x 6,027} \\
    \bottomrule
  \end{tabular}
\end{table}

To evaluate the effect of non-native multi-accent fine-tuning on native English speech, we also use the \textit{test-clean} subset of LibriSpeech \cite{panayotov2015librispeech}, a well-known ASR corpus made of recordings of native English speakers who read books, for testing purposes only. 
This subset contains 5~h 48~min of audio data.

\subsection{General parameters}
Our experiments were carried out using the Whisper small model \cite{radford2022robust}, which has encoder-decoder architecture and is of reasonable size (244M parameters) to run on many types of devices, such as on-board devices, and has proven to be an already highly capable ASR model.
It can be found on \textit{Hugging Face}\footnote{\url{https://huggingface.co/openai/whisper-small}}.
The fact that this model was trained on multilingual data is an important feature as multilingual features have proven to be useful to improve accented ASR \cite{matassoni2018non,vu2014improving}.
In order to match Whisper's expected input, all audio files have been resampled from 44.1~kHz to 16~kHz.
Models are trained for 3 epochs, with a batch size of 16.
Parameter-efficient fine-tunings were made by applying LoRA and \methodsname{} to attention modules in the Query and Value matrices (LoRA-qv and \methodsname{}-qv) or the Query, Value, Key and Output matrices (LoRA-qkvo and \methodsname{}-qkvo), where $r$ was set to 16 and $\alpha$ to 1.
These settings were chosen because they have proven to be effective \cite{hu2022lora, li2024mixlora}.
The learning rate is set to start at 1e-5 for full fine-tuning and 5e-5 for parameter-efficient fine-tuning methods, and decreases linearly to its half throughout the fine-tuning.
Fine-tunings have been conducted on NVIDIA A100 GPUs and tests on NVIDIA V100 GPUs.
For decoding, greedy search is used for computational reasons. 
Our code is publicly available\footnote{\url{https://gitlab.inria.fr/rbagat/mas-lora}}.

\subsection{Evaluation metric}
The results are reported in terms of the Word Error Rate (WER). 
Early stopping
is made using the WER on the validation set.
The statistical significance of the results has been validated using the Matched Pair Sentence Segment test with SCTK \cite{sctk}.

\begin{table*}[th]
  \caption{Zero-shot WER (\%) on test accents unseen during training. AR, ZH, HI, KR, SP and VI mean Arabic, Chinese, Hindi, Korean, Spanish, and Vietnamese accents, respectively. Bold numbers indicate the best result for each accent and those results which are statistically equivalent to it.}
  \label{tab:zeroshot}
  \centering
  \begin{tabular}{ c c | c c c c c c | c}
    \toprule
    {\textbf{Encoder}} & {\textbf{Decoder}} & {\textbf{AR}} & {\textbf{ZH}} & {\textbf{HI}} & {\textbf{KR}} & {\textbf{SP}} & {\textbf{VI}} & {\textbf{Mean}} \\
    \midrule
    No FT     & No FT &  13.18   & 16.04 & \textbf{7.64} & 10.40 & 13.72 & 22.01 & 13.77 \\
    Full FT     & Full FT &  15.50   & 20.89 & 11.06 & 14.72 & 17.11 & 23.42 & 17.12 \\
    \midrule
    LoRA-qkvo      & LoRA-qkvo                        &  \textbf{11.44}   & \textbf{15.70} & \textbf{7.36} & \textbf{9.54} & \textbf{12.46} & \textbf{19.80} & \textbf{12.72} \\
    \methodsname{}-qkvo  & LoRA-qkvo  & \textbf{11.43}  & \textbf{14.96} & \textbf{7.19} & \textbf{8.65} & \textbf{12.64} & \textbf{20.41} & \textbf{12.55} \\
    \bottomrule
  \end{tabular}
\end{table*}


\section{Results and discussions} \label{sec:results_discussions}
\subsection{Baselines}
We consider three baselines: pre-trained model without fine-tuning (referred to as No FT), full model fine-tuning (Full FT) and parameter-efficient fine-tuning using LoRA-qv or LoRA-qkvo. 
The obtained WERs can be found in rows 1, 2, 3 and 7 of Table \ref{tab:results_big}, respectively. 
It can be seen that full fine-tuning improves the performance compared to No FT.
LoRA applied to the Q, K, V, O matrices shows performance equivalent to No FT, while LoRA-qv shows significant improvements in performance compared to No FT, getting a WER of 12.32\%.

\subsection{Accent-agnostic \methodsname{}}
\textbf{Impact of \methodsname{} in the encoder} ---
Accent-agnostic \methodsname{} was studied under 3 conditions. 
As previously stated, it is always applied to the encoder.
For the decoder, we either used no fine-tuning or applied LoRA or \methodsname{}.
This allows us to see the effect of accent-related fine-tuning on the decoder which, we believe, should contain less accent-related features.
Results in Table \ref{tab:results_big} (rows 4-6, 8-10) show that \methodsname{}-qkvo significantly outperforms LoRA when applied to the encoder with LoRA-qkvo in the decoder, achieving a WER of 11.77\% compared to 13.48\% for LoRA-qkvo alone. 
For the Q, V matrices, LoRA-qv applied to the encoder and decoder yields a WER of 12.32\% and \methodsname{}-qv paired with LoRA-qv in the decoder 11.77\%.
\methodsname{} also significantly outperforms full fine-tuning, when applied to the encoder with LoRA in the decoder on both sets of matrices (11.77\% versus 12.21\%).



\textbf{Impact of \methodsname{} in the decoder}\label{sec:where_to_apply} ---
When \methodsname{} is used in the encoder, the results obtained by applying LoRA or \methodsname{} to the decoder (rows 5-6, 9-10) are similar to each other, with WERs of 11.77\% and 11.90\%, respectively. 
This indicates that accent-related fine-tuning isn't necessarily the best choice for the decoder. 
Instead, using an accent-independent method, here LoRA, is as effective.
Though, it is important to note that when \methodsname{} is used in the encoder, the decoder has to be fine-tuned. 
Not fine-tuning the decoder degrades the results, especially when \methodsname{}-qv is used in the encoder.

\textbf{Performance on native speech} ---
After fine-tuning the models on non-native speech, we tested them on native speech to evaluate the extent of performance degradation. 
The results are shown in the last column of Table \ref{tab:results_big}.
It can be seen that, compared to LoRA and Full FT, \methodsname{} yields results that are equivalent to those of the model before fine-tuning (rows 4-6, 8-9), except when \methodsname{}-qkvo is both applied to the encoder and the decoder.
This shows that \methodsname{} is less prone to catastrophic forgetting unlike full fine-tuning and LoRA.
In the following sections, when \methodsname{} is applied, we therefore use \methodsname{} in the encoder and LoRA in the decoder.

\textbf{Performance on unseen accents} ---
To assess the robustness of the method against new accents, we have conducted a zero-shot experiment by removing one accent from the training set and testing on that accent.
This was conducted for Full FT, LoRA-qkvo and \methodsname{}-qkvo.
According to Table~\ref{tab:zeroshot}, Full FT shows performance degradation on unseen accents compared to No FT.
On the other end, \methodsname{} remains as robust as LoRA in front of new accents.
Moreover, except in the case of the Hindi accent, both \methodsname{} and LoRA achieve significantly improved results compared to No FT, highlighting that multi-accent fine-tuning is important even if the training data does not cover test accents.

\subsection{Accent-aware \methodsname{}}
\begin{figure}[h]
\includegraphics[]{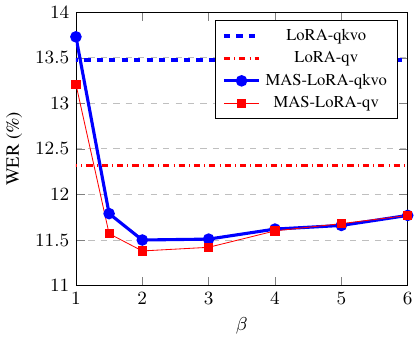}
\caption{Effect of $\beta$ on the WER when the accent label is known at inference. LoRA is applied to both encoder and decoder and \methodsname{} is applied to the encoder with LoRA in the decoder.}
\label{fig:beta}
\end{figure}

\textbf{Using only the specialized expert} ---
Accent-aware inference indicates that the accent label is known at inference.
One could then think that instead of using all the experts at inference, it would be better to use only the expert specialized in the 
sample's accent.
The results can be found in Fig.~\ref{fig:beta}, where $\beta=6$ indicates that all experts get equal weights and $\beta=1$ that only the expert specialized in the sample's accent is used.
It can be seen that using only 1 expert degrades the results compared to using all the experts.
This demonstrates the importance of combining knowledge from all experts, and can be interpreted as a form of regularization.

\textbf{Effect of $\bm\beta$} ---
The results obtained using accent-aware inference are shown in Fig.~\ref{fig:beta}.
The values of $\beta=1$ or 6 were already discussed in the previous paragraph.
For both \methodsname{}-qv and \methodsname{}-qkvo using $\beta=5$ already gives significantly better results than $\beta=6$, and performance keeps getting better as $\beta$ decreases until $\beta=2$. 
Though, as it can be seen, decreasing $\beta$ further has a negative effect on the WER.
This means that each expert has to contribute enough for \methodsname{} to be effective.
\section{Conclusion} \label{sec:conclusion}
In this article, we focused on the task of improving ASR when facing multiple non-native accents.
We introduced \fullmethodsname{}, a fine-tuning method based on a mixture of LoRA experts.
Each expert specializes in a specific accent, and their combined knowledge is used at inference.
We showed that, when the accent is unknown at inference, \methodsname{} significantly improves the WER compared to full fine-tuning and regular LoRA, provided that it is used in the encoder at least.
\methodsname{} also shows a similar generalization capability as LoRA when facing new accents and avoids catastrophic forgetting issues.
Moreover, when the accent is known at inference, \methodsname{} obtains further improved results.

\section{Acknowledgments}
This work was funded by the DeepMAUVES project supported by DGA and CNRS, and granted access to the HPC resources of IDRIS under the allocation 2024-AD011015024 made by GENCI.

\bibliographystyle{IEEEtran}
\bibliography{mybib}

\end{document}